\documentclass{article}

\usepackage{arxiv}

\usepackage[utf8]{inputenc} 
\usepackage[T1]{fontenc}    
\usepackage{hyperref}       
\usepackage{url}            
\usepackage{booktabs}       
\usepackage{amsfonts}       
\usepackage{nicefrac}       
\usepackage{microtype}      
\usepackage{lipsum}
\usepackage{epsfig}

\title{3D Shape Synthesis for Conceptual Design and Optimization Using Variational Autoencoders}

\author{Wentai Zhang \qquad Zhangsihao Yang \qquad Haoliang Jiang \qquad Suyash Nigam \\ \qquad \textbf{Soji Yamakawa} \qquad \textbf{Tomotake Furuhata} \qquad \textbf{Kenji Shimada} \qquad \textbf{Levent Burak Kara}\thanks{Address all correspondences to lkara@cmu.edu} \\ Carnegie Mellon University, Pittsburgh, PA, USA}

\begin{document}
\maketitle

\begin{abstract}
We propose a data-driven  3D shape design method that can learn a generative model from a corpus of existing designs, and use this model to produce a wide range of new designs. The approach learns an encoding of the samples in the training corpus using an unsupervised variational autoencoder-decoder architecture, without the need for an explicit parametric representation of the original designs. To facilitate the generation of smooth final surfaces, we develop a 3D shape representation based on a distance transformation of the original 3D data, rather than using the commonly utilized binary voxel representation. Once established, the generator maps the latent space representations to the high-dimensional distance transformation fields, which are then automatically surfaced to produce 3D representations amenable to physics simulations or other objective function evaluation modules. We demonstrate our approach for the computational design of gliders that are optimized to attain prescribed performance scores. Our results show that when combined with genetic optimization, the proposed approach can generate a rich set of candidate concept designs that achieve prescribed functional goals, even when the original dataset has only a few or no solutions that achieve these goals.
\end{abstract}


\section{Introduction}

In engineering design, while design simulation and analysis technologies are well developed and ubiquitous, digital tools to assist the early conceptual design phases are severely limited. Instead, humans still play a critical role in establishing the design space and the associated parameterizations. However, the heavy reliance on human-driven concept generation and design space exploration makes product development particularly challenging  for problems in which the geometry/form of the product has a significant impact on performance. As such, the need for digital design tools that support (1) knowledge extraction from configurationally and geometrically different past designs, (2) leveraging this information for large-variance, automatic design synthesis inside and outside of the original design space, and (3) seamless integration into analysis and simulation engines remain a central need in design automation.

In this work, we present a data-driven 3D shape synthesis method to assist human designers in  conceptual design. Our approach relies on the observation that past designs may encapsulate useful design information that, if digitally captured, could be used to generate new designs automatically. To this end, we adopt an unsupervised variational autoencoder (VAE)  deep learning method that takes input a corpus of 3D designs and extracts a latent design representation. This representation transforms the originally very high-dimensional data  into a compact feature vector, where each feature encodes a latent probability distribution function learned over all past designs. Once learned, this representation can be sampled, or elements in this latent space can be interpolated and extrapolated to generate new latent space instances. These new instances can then be projected to the original design space using the decoder of the VAE.  

In contrast to the common method of voxelizing 3D data into a 3D binary representation \cite{DBLP:conf/nips/LiPSQG16,3dshapenet,DBLP:conf/bmvc/SedaghatZAB17,DBLP:journals/corr/BrockLRW16}, we utilize distance transformation maps \cite{DBLP:conf/cvpr/StutzG18} as the primary input. This involves (1) the conversion of input 3D shapes (commonly acquired in the form of polygonal models) into real valued distance maps, (2) using the distance map of each original design to train the VAE, (3) automatically converting any synthetically generated new distance map back to a polygonal model for downstream analysis. This allows the synthesized  3D shapes to exhibit much smoother surfaces without suffering from `pixelatization,' while being amenable to  engineering analyses.   

We demonstrate the utility of our approach on the 3D outer shape design of gliders. While one approach is to learn a direct mapping from the available past designs to their aerodynamic performance, and use this mapping as a simulator to evaluate new designs, such a mapping would need to be learned for every new engineering objective. Instead, VAE learns a shape generator in an unsupervised manner, where the latent space exploration in the trained architecture allows the newly generated design to be integrated into widely available analysis tools. We demonstrate this on two particular case studies, where the first study involves a purely geometric assessment of the objective while the other incorporates flight dynamics (albeit simplified) for shape optimization. 

Our approach also develops a simple latent space design crossover technique that allows a genetic optimizer to produce a large set of new designs through stochastic latent vector interpolation and extrapolation. While not a requirement in the overall framework, the use of genetic optimization enables a large set of synthetically generated designs to meet the target performance score, even though only a few or no original design solutions could attain the prescribed goals. This, in turn, offers greater conceptual latitude to the human designer in deciding which concepts to further develop.

\begin{figure*}[tp]
  \centering
  \includegraphics[trim = 0in 0in 0in 0in, clip, width=\textwidth]{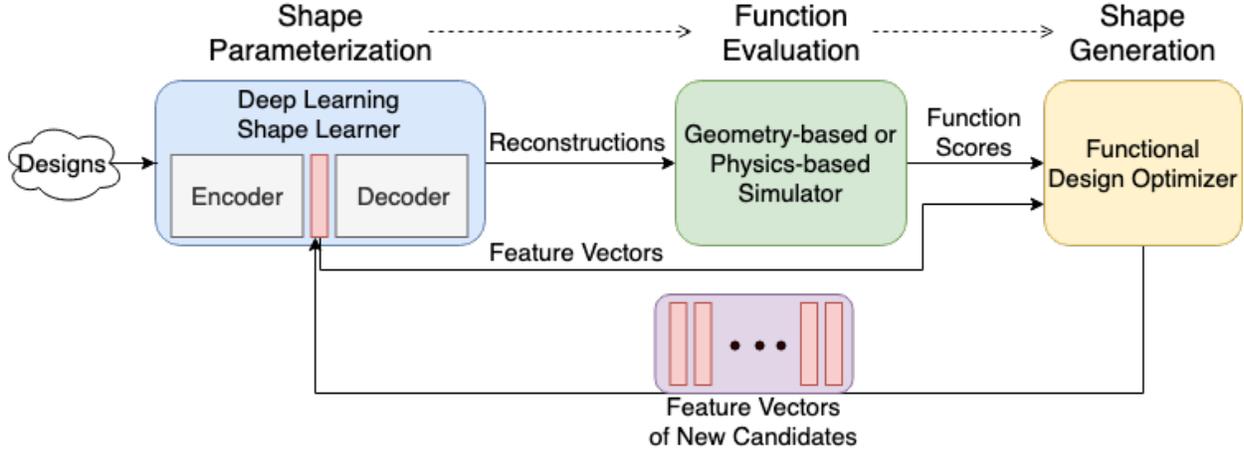}
  \caption{The architecture of our proposed data-driven conceptual design pipeline.}
  \label{fig:pipeline}
\end{figure*}

\section{Background}

With advances in machine learning, data-driven approaches that model and optimize engineering design problems are becoming increasingly more prevalent. Sosnovik et al. \cite{DBLP:journals/corr/abs-1709-09578} propose a Convolutional Neural Network that accelerates  topology optimization computations. Using the powerful ability of deep learning methods to segment images pixel-wise, the approach predicts final optimal topologies after several iterations of optimization, based upon the initial conditions of the layout. 

To address the circuit synthesis in EDA design, Guo et al. \cite{guo2018eda} propose an active learning strategy for reducing topology evaluation cost for a circuit synthesis problem. They utilize a predictive model with a random forest to approximate true circuit topology performance. Their experimentation reveals that uncertainty and topology structure may play critical roles in improving the appropriation model accuracy and make a significant contribution to reducing the system evaluation costs.

Data-driven methods have also been used in modeling non-linear physics. Raissi et al. \cite{raissi2018nonlinear} treats deep neural network as a non-linear function approximator and use the method to identify complex non-linear systems such as Lorenz system and the Glycoltic oscillator model. Umetani et al. \cite{Umetani:2014:PID:2601097.2601129} develop a data-driven approach to estimate the aerodynamic forces on a glider and its wing shape, and use this for glider design. This enables a user to accurately match the desired trajectory without the aid of costly simulations or experiments.

With large amount of available data and the advance of hardware technology, data-driven methods have become an increasingly common strategy for the problems that are difficult to approach by creating physical model or are expensive in computation. Recently, researchers in the mechanical design community have started exploring machine learning approach to aid the design process. 

Fuge et al. \cite{fuge:jmd_2014_hcd_recommendation} devise a framework which relies on collaborative filtering to recommend best design methodologies to solve target design problems, and argue that such  approaches can be valuable for novice designers and enhance the overall product development cycle.
In order to have an automatic design generator, Chen et al. \cite{DBLP:journals/corr/abs-1808-08871} introduce the B$'$ezierGAN, a generative model for synthesizing smooth curves. The model maps a low-dimensional latent representation to a sequence of discrete points sampled from a rational B$'$ezier curve. It is tested on four different design datasets and reveals better capacities in generating realistic 2D smooth shapes when compared with InfoGAN. Similar frameworks can have an impressive performance even for creative hand sketches. Chen et al. \cite{DBLP:journals/corr/abs-1709-04121} propose a model, sketch-pix2seq, based on a sequence-to-sequence variational-auto-encoder (VAE) model called sketch-rnn.  With their modification, the model has better performance in learning and generating sketches of multiple categories and shows promising results in creative tasks. However, these works are limited to 2D images or sketches. It is still challenging to extend these works to 3D design representations.

To have a deeper understanding of the mapping between the shape and the functions of design, Dering et al. \cite{dering2017convolutional} propose a deep learning approach based on three-dimensional (3D) convolutions that predict functional quantities of digital design concepts. Testing trained models on novel input yields accuracy as high as 98\% for estimating rank of the functional quantities. This method is also employed to differentiate between decorative and functional headwear. Moreover, Burnap et al. \cite{2018arXiv181211067B} develop a deep learning approach to predict design gaps in the market. Their approach is built on conventions in both quantitative marketing inbounding the heterogeneity of consumer choice preferences, as well as engineering design for bounding the space of possible designs. Raina et al. \cite{raina2018design} explore the representation of design strategies as a Hidden Markov Model and their application to engineering design problems. Their results imply the successful transfer of design strategies from human designers to computational agents. They also propose a method to achieve transfer learning in agent-based models through state-based probabilistic models. Burnap et al. \cite{Burnap2016EstimatingAE} uses a deep learning based generative model to find the statistical representation of a design space via using large number of images and design attributes. They test their methods on automobile body design and successfully morph a vehicle into different meaningful body types. In consideration of a sequential design pipeline, Oh et al.  \cite{oh2018design} manage to combine the generative methods with further topology optimization in automobile wheel design. However, the synthetic designs are still technically immature with one-shot optimization. They claim that an iterative and automatic optimization process can be a better alternative.

Although the aforementioned prior researches achieve plausible results when applying data-driven approach as a tool for a single design process, there doesn't exist a close-loop conceptual design pipeline which enhances the design automation and optimization with respect to the functional requirements.

\section{Technical Approach}

 Fig.~\ref{fig:pipeline} shows our proposed framework. The key modules are a deep learned  shape encoder-decoder, a geometry or physics-based design simulator, and an optimizer. 
 
Input to our approach is a database of 3D models belonging to the same object category (e.g., aircrafts). These models are most commonly acquired in the form of 3D polygonal surface models. Through unsupervised learning, the variational shape learner extracts a latent feature vector for each of the input designs. The latent space vector has much fewer dimensions than the original shape representation, and thus serves as a dimension reduced encoding of the large design space. 

The geometry- or physics-based simulator is determined by the design performance objectives. It is responsible for testing the design candidates and provide performance scores for the subsequent optimization process. 

The optimizer utilizes the outcomes of the simulator, together with the latent space representation, to optimize the designs directly in the latent space.  These modules are detailed further in the following sections.

\subsection*{Variational Shape Learner}

\begin{figure*}[tp]
  \centering
  \includegraphics[trim = 0in 0in 0in 0in, clip, width=\textwidth]{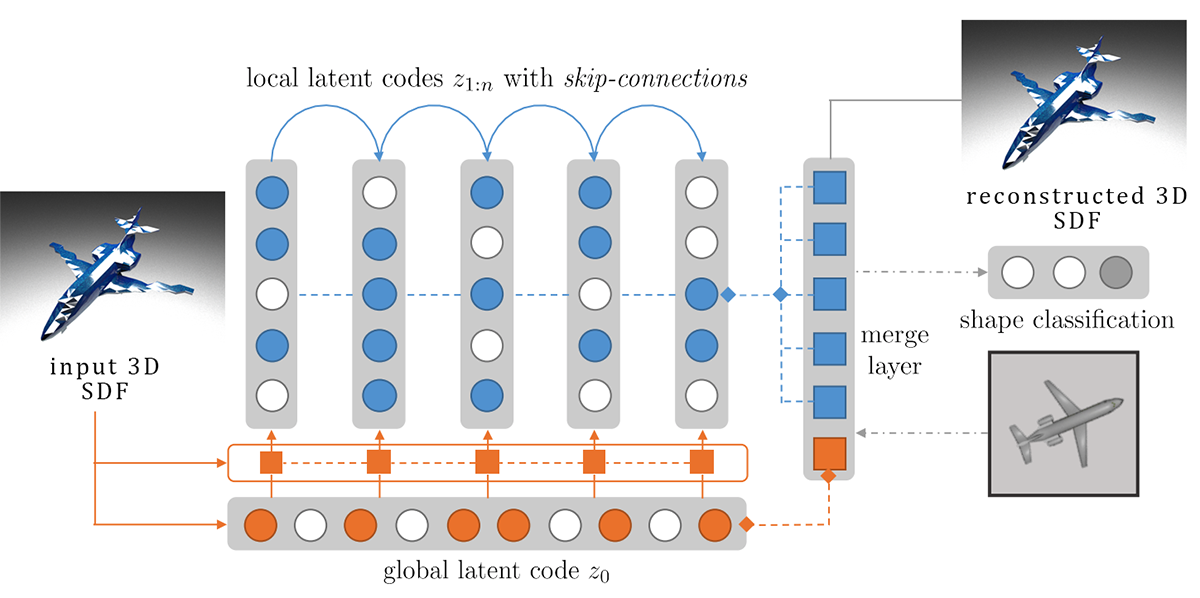}
  \caption{A schematic  of VSL\cite{DBLP:conf/3dim/LiuGO18}.}
  \label{fig:vsl}
\end{figure*}

In this study, we adopt a variational autoencoder model, the Variational Shape Learner (VSL) \cite{DBLP:conf/3dim/LiuGO18}, which builds on the ideas of the Neural Statistician \cite{edwards2016towards} and the volumetric convolutional network \cite{maturana2015voxnet}. The parameters of the VSL are learned under a variational inference scheme \cite{kingma2013auto}. As shown in Fig.~\ref{fig:vsl}, we use a hierarchical VAE (Variational Autoencoder) which consists of an encoder, a decoder and a latent space feature representation.

\textbf{Distance Maps and Shape Representation}: We use a 3D signed distance field (SDF) as the primary design representation in the original space. As opposed to the commonly used binary voxel representation  \cite{DBLP:conf/nips/LiPSQG16,3dshapenet,DBLP:conf/bmvc/SedaghatZAB17,DBLP:journals/corr/BrockLRW16}, this representation allows smooth final surfaces to be constructed over the designs generated by VSL.

SDF is a scalar function of position that defines a closed volume implicitly.  The absolute value of the function is the distance from the surface of the solid.  A positive value indicates the point is inside of the solid, or negative outside.  The boundary is the isosurface where the function value is zero. SDF has been recently used for deep-learned shape completions  \cite{DBLP:conf/cvpr/StutzG18}.

In our approach, the SDF is implemented as a tri-linear function defined over a structured lattice.  A signed distance is assigned to the lattice nodes and interpolated within the lattice cell.  A lattice with $n\times m \times k$ nodes has $(n-1) \times (m-1) \times(k-1)$ cells, and each node is connected to neighboring nodes by multiple edges.

To create the training and testing data, an SDF needs to be generated from a polygonal mesh.  The program calculates a distance $d$ from each lattice node to the nearest point on the polygonal mesh.  Then $+d$ is assigned to the node if the node is inside of the solid, or $-d$ if outside.

\begin{figure}[h]
  \centering
  \includegraphics[trim = 0in 0in 0in 0in, clip, scale=0.8]{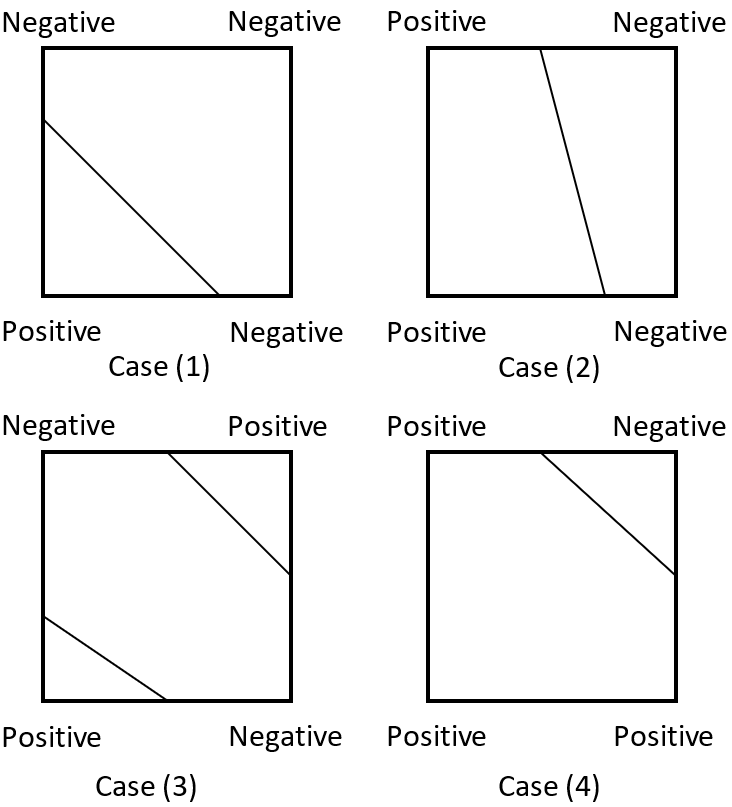}
  \caption{Four cases of arc generation in a rectangular face of a lattice cell.}
  \label{fig:sdf}
\end{figure}

Once an SDF is obtained as an output from the VSL, a polygonal mesh of the boundary needs to be extracted from the SDF for rendering and to facilitate downstream processes.

For this, our approach first places a vertex on each edge with the two nodes having one positive and one negative value.  The position of the vertex is smoothly interpolated based on the distance values.

Then the program creates arcs on each rectangle of the lattice cells that has mix of positive and negative nodes.  There are four cases: (1) one negative and three positive, (2) two negative connected and two positive connected, (3) two negative diagonal and the other two positive, and (4) one positive and three negative.  Cases (1), (2), and (4) yield one arc, and (3) yields two arcs as shown in Fig.~\ref{fig:sdf}.

Finally, polygonal faces in each lattice cell are created by connecting the arcs. The result may become degenerate if a lattice node has exactly zero value, or the node is exactly on the boundary.  We append a very small value to the distance values of such nodes to avoid this degeneracy.

\textbf{Encoder-Decoder Details}: For the encoder, the global latent code is directly learned from the input SDF through three convolutional layers with kernel sizes {6, 5, 4}, strides {2, 2, 1} and channels {32, 64, 128}. Each layer is followed by a ReLU activation layer \cite{nair2010rectified} and a batch normalization layer \cite{DBLP:conf/icml/IoffeS15}. Each local latent code is conditioned on the global latent code, the input voxel and the previous latent code (except for the first local latent code) using two fully-connected layers with 100 neurons each. After we learn the global and local latent codes, we concatenate them into a single vector. A 3D deconvolutional neural network with dimensions symmetrical to the encoder of the global latent code is used to decode the learned latent features into an output SDF model. 

An element-wise logistic sigmoid \cite{Mitchell:1997:ML:541177} is applied to the output layer in order to convert the learned features to occupancy probabilities for each SDF cell. The detailed network architecture can be found in  \cite{DBLP:conf/3dim/LiuGO18}.

\textbf{Data}: For the data, we use the 4096 airplane models from the ModelNet40 repository \cite{3dshapenet}.  We  clean the floating patches in the models and   align all the models in the same direction. Then we compute the SDF from the polygonal model  to produce the network input  $[41\times 41\times 41]$ in size for each model. It takes 3 hours to train the network for 1000 epochs on an NVIDIA GeForce 1080 GPU. The learning rate is $5\times 10^{-3}$, and the batch size is 64.

\subsection*{Physics-based Simulator}

The trajectories of the original and synthesized aircrafts are simulated in YS FLIGHT SIMULATOR \cite{ysflight}. While the flight dynamics kernel of the simulator uses a  simplified physical model, it is  suitable for making a quick  estimation of the flight characteristics.

An aircraft in the air is subject to lift, drag, gravity, and thrust, so called the four forces of flight.  For this design task thrust is cut to idle, or zero, rendering the designs as gliders.  Gravity is -9.8$m/s^2$ in the y-direction.  Lift and drag forces are calculated as:
\begin{equation}
L=\frac{1}{2} C_L \rho v^2 S
\end{equation}

\begin{equation}
D=\frac{1}{2} C_D \rho v^2 S,
\end{equation}

\noindent where $L$ is lift, $C_L$ is lift coefficient, $\rho$ is air density, $v$ is velocity, $S$ is wing area, $D$ is drag, and $C_D$ is drag coefficient.  $C_L$ and $C_D$ are functions of $\alpha$ or angle-of-attack.  YS FLIGHT SIMULATOR kernel approximates $C_L$ and $C_D$  as a piecewise-linear and a parabolic functions respectively.

The rotation of an aircraft is the hardest to simulate.  Unless the moment of inertia and center of gravity is known it is impossible to make an accurate simulation. Instead of estimating the moment of inertia and center of gravity, the simulator kernel approximates a rotation as a second-order system with a stability constant and a maneuverability constant, both of which are empirically specified.

Estimating a reasonable $C_L$ and $C_D$ functions and stability and maneuverability constants for an output design of the network is also a challenge.  However, the output design from the  network leaves freedom of choosing the airfoil of the wing, moment of inertia, and the center of gravity.  It is reasonable to assume that the designs generated by the  network can be configured to have a similar characteristic to an existing airplane of a similar geometric signature.  We have chosen one from 88 aircraft data in the flight simulator that matches the forward-projection and top-view-projection area ratio the best to the output design from the  network and take $C_L$ and $C_D$ functions and stability and maneuverability constants for the characteristic of the designs generated by the  network.

\begin{figure}[!tp]
  \centering
  \includegraphics[trim = 0in 0in 0in 0in, clip, width=\columnwidth]{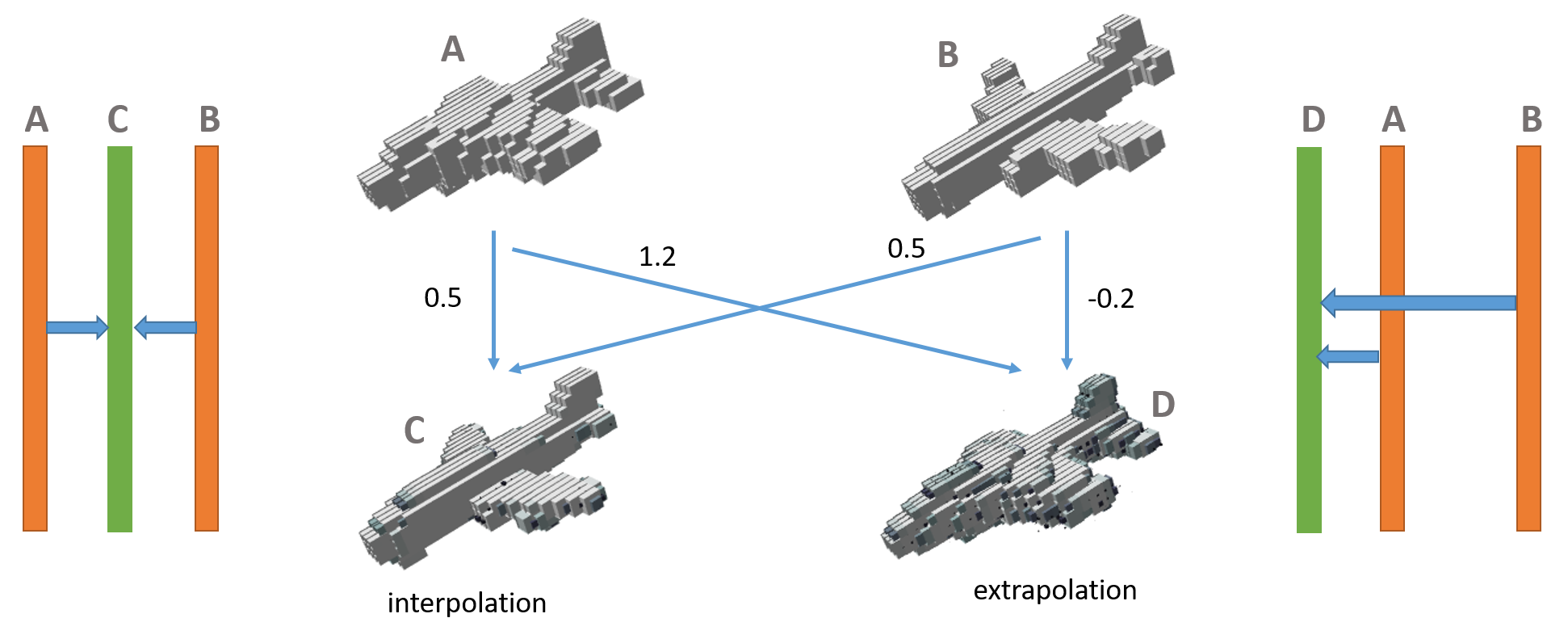}
  \caption{Sample reconstructions of the latent vector interpolation and extrapolation.}
  \label{fig:inter}
\end{figure}

\subsection*{Design Optimizer}

The main purpose of this module is to explore potentially better design candidates using the compact latent space. We use a gradient free genetic optimizer following its success in conceptual design \cite{GA_enlargement,GA_Conceptual,GADesignSpaceSearch}. Latent feature vectors extracted from various designs are selected and ranked  based on their corresponding performance scores. Then, mutation and crossover operations involving interpolation and extrapolation are used to produce subsequent generations.

We observe that conventional crossover renders  shapes incomplete. To address this problem, we utilize the line crossover operator \cite{lineCrossover} to generate new offsprings. A child is generated using a linear interpolation between two parents:
 \begin{equation}
Child = r \times P_1 + (1 - r) \times P_2,
\end{equation}

\noindent where $r\in[0,1.2]$ is a random number drawn from a uniform distribution, and $P_1$ and $P_2$ are the two parents respectively. The intuition of line crossover is that the linear interpolation or extrapolation of the two parents may provide competitive offsprings \cite{GADesignSpaceSearch}. In our case, we expect the line crossover is functionally similar to shape morphing which blends the geometric representation  of two models together. The weight $r$ tells the similarity between the child representation and the parent $P_1$. Sample results of the reconstruction models from the latent feature interpolation and extrapolation are shown in Fig.~\ref{fig:inter}. Note that  the voxelized representation is used to show the coarse, binary versions of the designs that are generated from the latent space. The actual output of the network is an SDF.

We first compute the scores for all the models in our dataset. Then, we rank all models and randomly select one model from every $(N/n)$ interval where $N$ is the number of total models, $n$ is the population size (set to 100 in our approach). The probability of crossover and mutation is 0.9 and 0.05 respectively.

The optimization goal is to minimize the Mean Square Error:
\begin{equation}
MSE = ||y_t-y_c||_2,
\end{equation}

\noindent where $y_t$ denotes the target functional objective score and $y_c$ denotes the scores of the children generated in each generation.

\section{Case Study}

In this section, we demonstrate our approach on glider aircraft design. All the modules in the pipeline are specified from the suggested options we mentioned above. Details of the design task and each module will be described in the following subsections.

\begin{figure}[!tp]
  \centering
  \includegraphics[trim = 0in 0in 0in 0in, clip, width=\columnwidth]{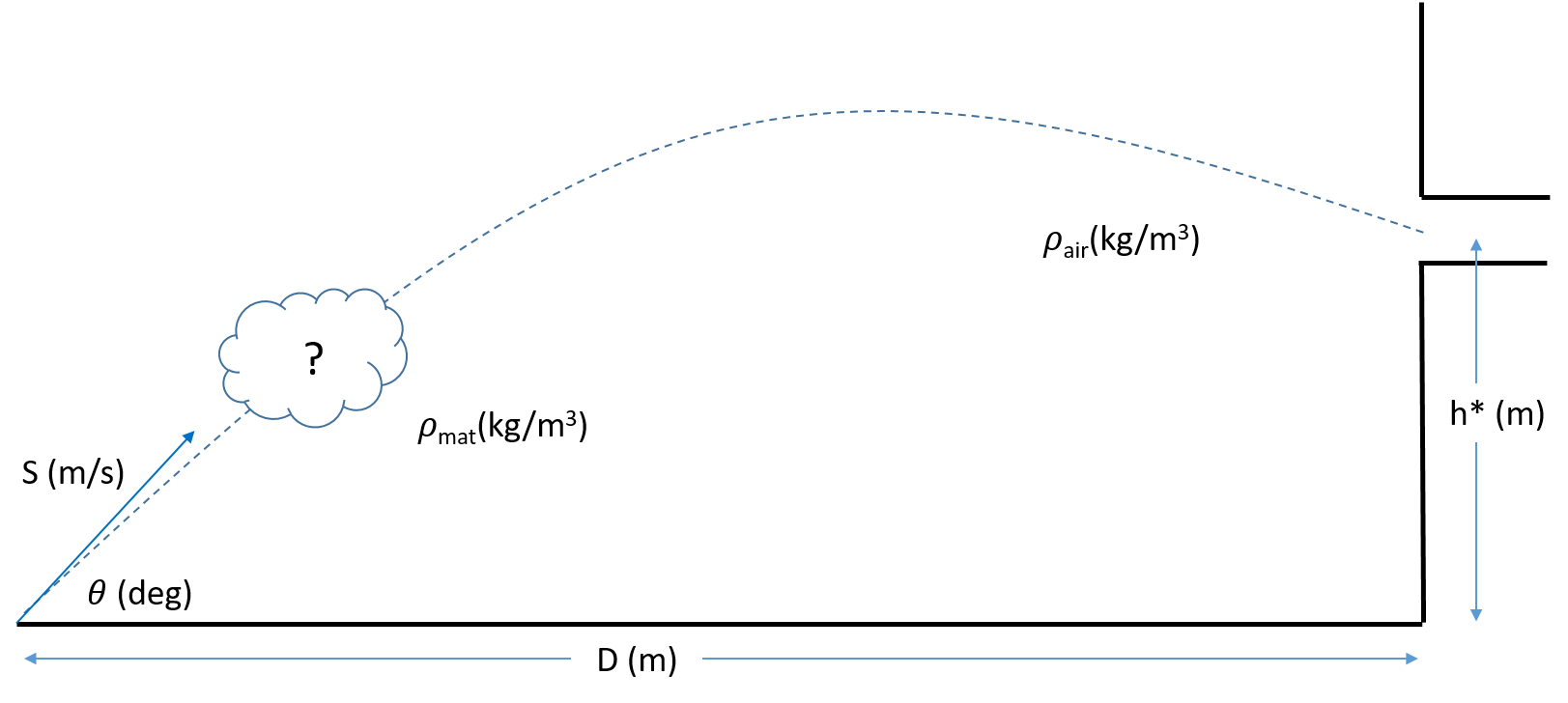}
  \caption{A schematic graph of the design task.}
  \label{fig:challenge}
\end{figure}

\subsection*{Design Task}
As shown in Fig.~\ref{fig:challenge}, given the initial launch speed $S(m/s)$, the initial pitch angle $\theta(degree)$, the density of the projectile aircraft $\rho_{mat}(kg/m^3)$ and the density of the air $\rho_{air}(kg/m^3)$ are given, the goal is to design the shape of the glider so that it can go through the gap located at  $h^*$m high from the ground, and $D$m horizontally away from the launch point.

In this design task, the constraint is that the projectile candidates should all fit in a $1m\times1m\times1m$ box without any propulsion. Among all the parameters, $D$, $\rho_{mat}$ and $\rho_{air}$ are fixed. $S$, $\theta$ as well as $h^*$ can be set arbitrarily by the user. Designers are usually accustomed to the inverse task: Given the projectile, adjust $S$ and $\theta$ to hit the target height, which  can be regarded as a tuning process of several parametric design variables. By contrast, in our design problem, the shape of the projectile aircraft cannot be readily parameterized or represented with a limited number of design variables.

\section{Results}
\begin{figure}[tp]
  \centering
  \includegraphics[trim = 0in 0in 0in 0in, clip, width=0.9\columnwidth]{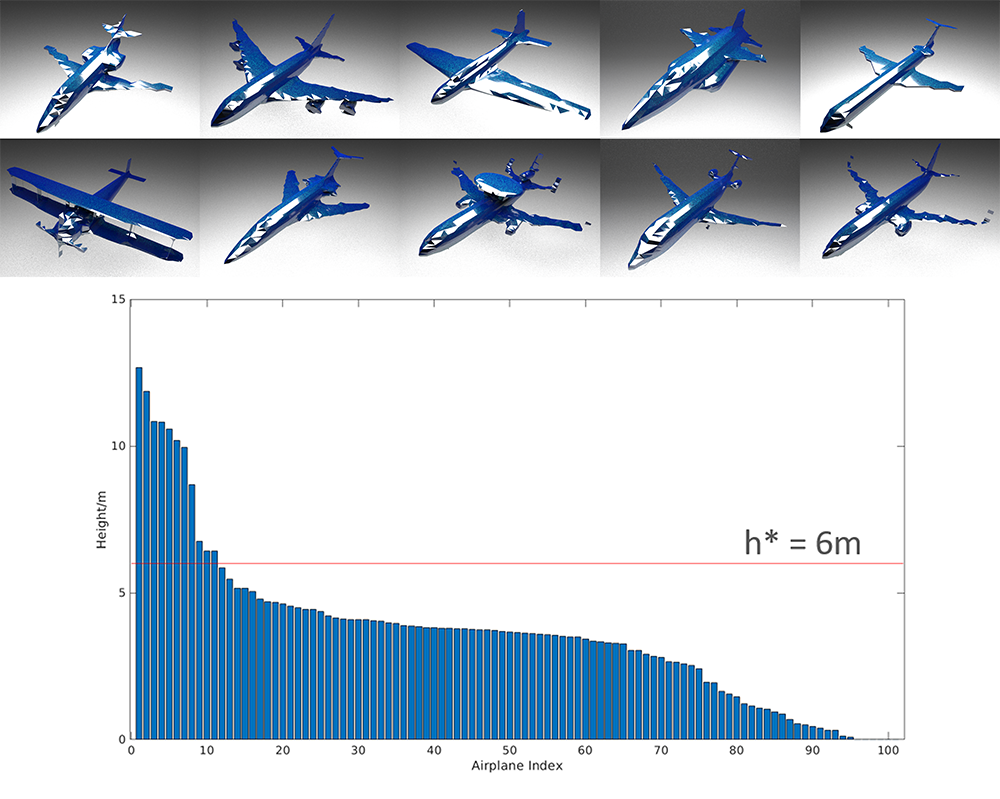}
  \caption{\textbf{Upper: }10 randomly selected aircraft SDF models from the initial population. \textbf{Lower: }The landing height $h (m)$ distribution of the 100 initial population.}
  \label{fig:initial}
\end{figure}

In our experiments, we set $\theta=10^o$, $S=45.7m/s$, $\rho_{mat}=1000 kg/m^3$, $\rho_{air} = 1.29 kg/m^3$ and $D=100m$. Our main interests lie in the height of the gap $h^*(m)$. We varies the value of $h^*$ to test the performance of our proposed pipeline. 

To have an intuitive sense of a reasonable range for $h*$, we randomly select 100 aircraft models from the dataset. Then we obtain the landing height of each aircraft model using the physics-based simulator we introduce in the CASE STUDY section. The acquired height range is $h\in [0, 12.6]m$. A plot of the sorted height distribution is shown in Fig.~\ref{fig:initial}. Note that zero means the model lands at the ground before $D=100m$. 

\begin{figure}[tbp]
  \centering
  \includegraphics[trim = 0in 0in 0in 0in, clip, width=0.9\columnwidth]{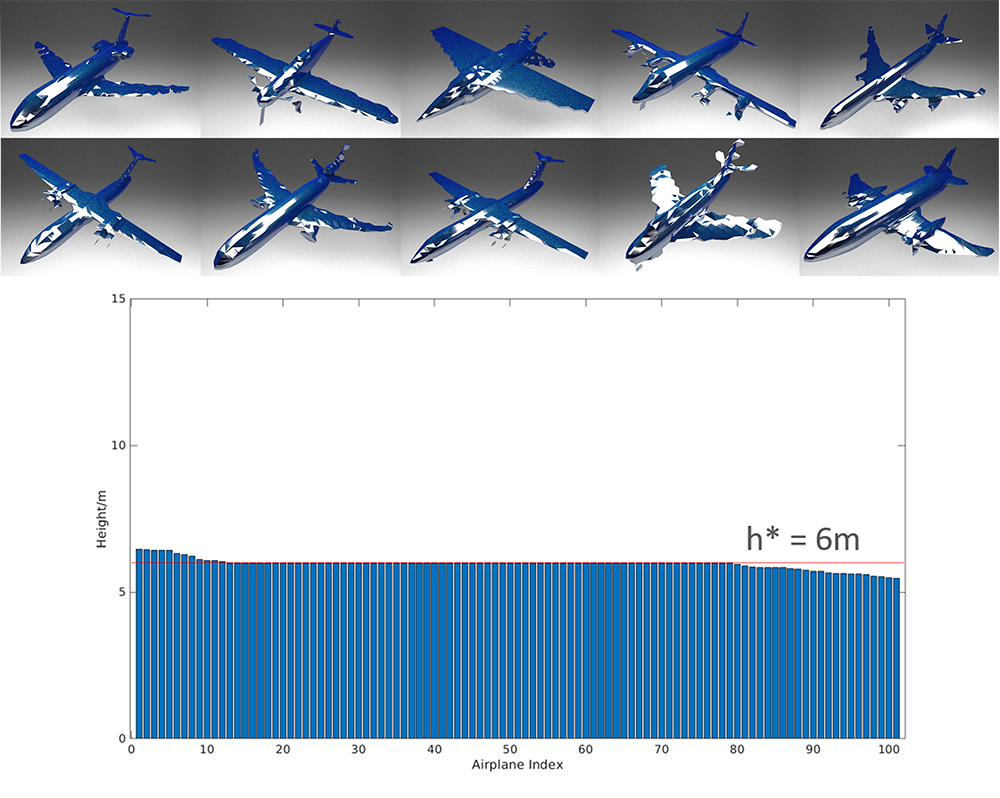}
  \caption{\textbf{Upper: }10 randomly selected aircraft SDF models from the optimized population when $h^* = 6 m$. \textbf{Lower: }The landing height $h (m)$ distribution of the 100 optimized population when $h^* = 6 m$.}
  \label{fig:h6}
\end{figure}

First, we set $h^*=6m$, which is around the mean value of the height range. After about 200 rounds of an iterative optimization process, the distribution of the landing heights for the final design candidates are demonstrated in Fig.~\ref{fig:h6}. Unlike the diverse distribution in Fig.~\ref{fig:initial}, the distribution is much flatter and mostly gather near the target height of $h^*=6m$. In fact, if the tolerance is 0.1 m, 76\% of design candidates in the final population satisfy the design requirement. (See Tab.~\ref{table_h6}) In the initial population, only 1\% of the population can fulfill the same requirements. Fig.~\ref{fig:h6} also shows the diversity in the final synthetic populations, which will benefit the human designers with more valid options to choose from.

\begin{figure}[tbp]
  \centering
  \includegraphics[trim = 0in 0in 0in 0in, clip, width=0.9\columnwidth]{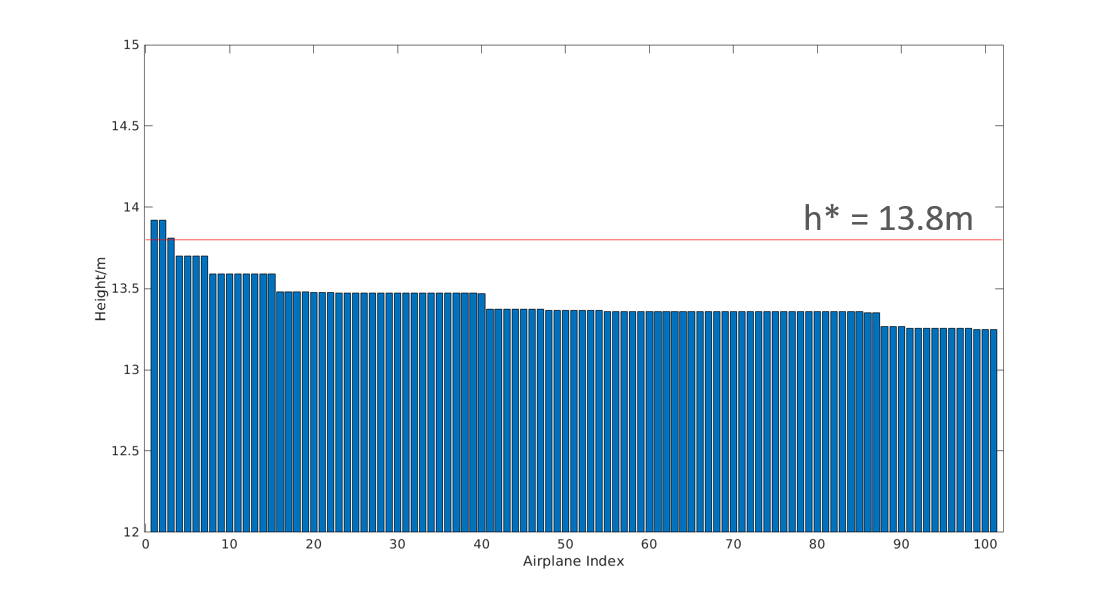}
  \caption{The landing height $h (m)$ distribution of the 100 optimized population when $h^* = 13.8 m$.}
  \label{fig:h138}
\end{figure}

\begin{figure}[tbp]
  \centering
  \includegraphics[trim = 0in 0in 0in 0in, clip, width=0.9\columnwidth]{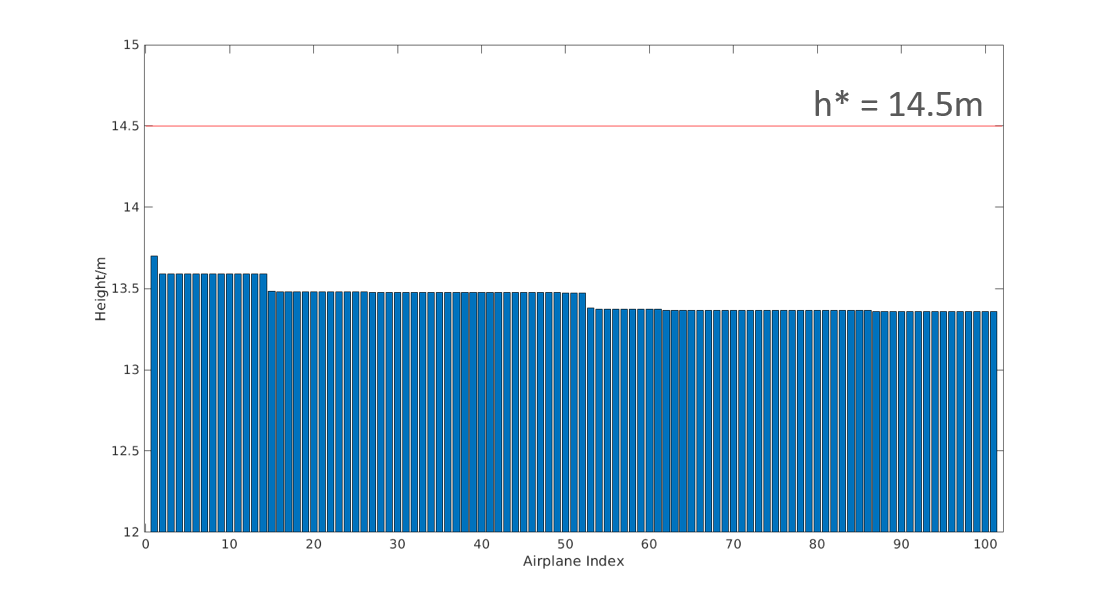}
  \caption{The landing height $h (m)$ distribution of the 100 optimized population when $h^* = 14.5 m$.}
  \label{fig:h145}
\end{figure}

To demonstrate the ability of our pipeline in design space exploration, we intentionally set a design requirement that is originally unfeasible in the existing design candidates. Specifically, we set $h^*=13.8m$. The height distribution of the suggested design candidates after optimization is shown in Fig.~\ref{fig:h138}. We can observe that all the heights are within $[13.2, 13.9]m$, which exceed the maximum height($12.6m$) the initial designs can reach. According to Tab.~\ref{table_h138}, although the concentration of heights for final designs is not as impressive as $h^*=6m$ case, 88\% of candidates are able to fulfill the requirement if the tolerance is $0.5m$. The results also reveal the difficulty of exploration outside the original feasible space.  To make the task even more challenging, we set $h^*=14.5m$ and repeat all the design process. Expectedly, see Fig.~\ref{fig:h145}, the final synthetic candidates cannot reach such a height requirement. But we obtain a design boundary reference in a data-driven manner as well as plentiful promising designs that provide useful suggestions for further manual design exploration.

\begin{table}[tbp]
\caption{PERCENTAGE OF THE DESIGN CANDIDATES WITH THE LANDING HEIGHT $h$, S.T. $|h-h^*|<\delta$ WHEN $h^*=6m$.}
\begin{center}
\label{table_h6}
\begin{tabular}{c l l}
& & \\ 
\hline
$\delta(m)$ & Initial & Final \\
\hline
0.1 & 1\% & 76\% \\
0.5 & 4\% & 98\% \\
\hline
\end{tabular}
\end{center}
\end{table}

\begin{table}[tbp]
\caption{PERCENTAGE OF THE DESIGN CANDIDATES WITH THE LANDING HEIGHT $h$, S.T. $|h-h^*|<\delta$ WHEN $h^*=13.8m$.}
\begin{center}
\label{table_h138}
\begin{tabular}{c l l}
& & \\ 
\hline
$\delta(m)$ & Initial & Final \\
\hline
0.1 & 0\% & 5\% \\
0.5 & 0\% & 88\% \\
\hline
\end{tabular}
\end{center}
\end{table}

\section{Conclusions}
This work presents an integrated conceptual design pipeline involving a data-driven shape learner, a function evaluator and a functional design optimizer. The pipeline is verified with a case study on the shape design of projectile aircraft models. When the design objective is set within the objective range of the design exemplars, our algorithm is able to synthesize a large set of design candidates which also satisfy the functional design requirements. It is capable of generating new designs whose performance scores are outside the range of the original models.  

Likewise, the same approach can be utilized to generate valid design candidates in other domains when adopting different options for each module. For example, if the design focus is the shape of a 2D beam bridge with load constraints, the shape learner alternative can be a 2D VAE. The physics-based simulator should accordingly be an FEA simulator. The functional design optimizer can adopt PSO. 

At the meantime, even for design problems with the same dimension (2D/3D), various data representations can be introduced in the pipeline in terms of the focus area. Depth images are suitable when designing shell structures. While point clouds are beneficial when designing adjacent mesh surfaces. These potential extensions of our pipeline are becoming increasingly promising as several recent learning frameworks like Matterport3D \cite{DBLP:conf/3dim/ChangDFHNSSZZ17} and PointNet \cite{DBLP:conf/cvpr/QiSMG17} have been developed.    

\section{Limitations and Future Work}
Two time-consuming processes in our pipeline are the training of the shape learner and the geometry-based or physics-based simulation. A potential more efficient method is to directly establish a mapping between the latent feature vector and its corresponding physic properties, which may be a trained network as well. In this manner, designers can completely get rid of the original data format (2D/3D) during the iterative optimization process. The final proposed designs will be reconstructed once the objective is fulfilled.

When manipulating the feature vectors in the functional optimizer module, we don't have a sense of the exact meaning of each dimension in the latent space. This is a common problem which has also problematic for the researchers from the machine learning field. Recently, researchers like Wieczorek et al. \cite{DBLP:journals/corr/abs-1804-06216} are exploring learning orthogonal latent features, which may enable the designers to acquire a desired latent space.

\newpage
\bibliographystyle{unsrt}  
\bibliography{references}  






\end{document}